\documentclass[10pt,twocolumn,letterpaper]{article}

\usepackage{cvpr}
\usepackage{times}
\usepackage{epsfig}
\usepackage{graphicx}
\usepackage{amsmath}
\usepackage{amssymb}
\usepackage{xcolor}
\usepackage{booktabs}
\usepackage{paralist}
\usepackage{bbm}
\usepackage{multirow}

\usepackage[pagebackref=true,breaklinks=true,letterpaper=true,colorlinks,bookmarks=false]{hyperref}

\newcommand*{\resp}{resp.\@\xspace}

\DeclareMathOperator*{\argmin}{\arg\!\min}

\newcommand{\opt}[1]{\tilde{#1}}
\newcommand{\gt}[1]{{#1}^*}

\newcommand{\pos}{\mathbf{p}}
\newcommand{\crd}{\mathbf{y}}

\newcommand{\mdl}{\mathbf{h}}
\newcommand{\trans}{\mathbf{t}}
\newcommand{\rot}{\boldsymbol{\theta}}

\newcommand{\refine}{\mathbf{R}}
\newcommand{\loss}{\ell}
\newcommand{\res}{\mathbf{r}}

\newcommand{\paramw}{\mathbf{w}}

\newcommand{\expectation}[2]{\mathbb{E}_{#1}\left[ #2 \right]}
\newcommand{\derv}[1]{\frac{\partial}{\partial #1}}

\newcommand\rgb{\mbox{RGB}\@\xspace}
\newcommand\rgbd{\mbox{RGB-D}\@\xspace}

\cvprfinalcopy 

\def\cvprPaperID{****}

\ifcvprfinal\fi
\begin{document}

\title{Learning Less is More -- 6D Camera Localization via 3D Surface Regression}

\author{Eric Brachmann and Carsten Rother\\
Visual Learning Lab\\
Heidelberg University (HCI/IWR)\\
{\tt\small http://vislearn.de}
}

\maketitle
\thispagestyle{empty}

\begin{abstract}
Popular research areas like autonomous driving and augmented reality have renewed the interest in image-based camera localization. 
In this work, we address the task of predicting the 6D camera pose from a single \rgb image in a given 3D environment. 
With the advent of neural networks, previous works have either learned the entire camera localization process, or multiple components of a camera localization pipeline. 
Our key contribution is to demonstrate and explain that learning a single component of this pipeline is sufficient. 
This component is a fully convolutional neural network for densely regressing so-called scene coordinates, defining the correspondence between the input image and the 3D scene space. 
The neural network is prepended to a new end-to-end trainable pipeline. 
Our system is efficient, highly accurate, robust in training, and exhibits outstanding generalization capabilities. 
It exceeds state-of-the-art consistently on indoor and outdoor datasets. 
Interestingly, our approach surpasses existing techniques even without utilizing a 3D model of the scene during training, since the network is able to discover 3D scene geometry automatically, solely from single-view constraints.
\end{abstract}

\begin{figure}
\begin{center}
\includegraphics[width=0.9\linewidth]{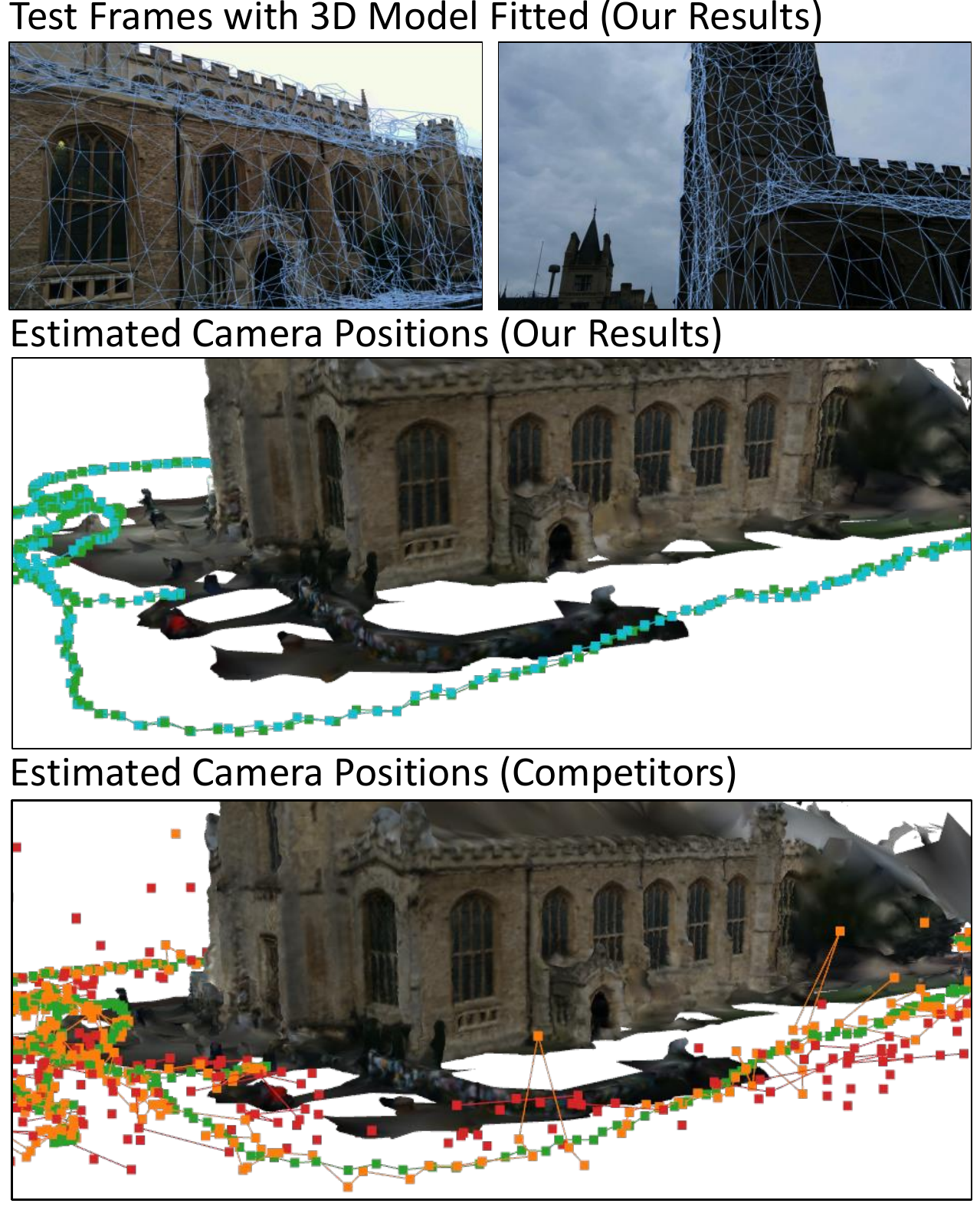}
\end{center}
   \caption{\textbf{One-Shot \rgb Camera Localization. Top:} 
   We use our pose estimates to fit a 3D scene model to test images.
\textbf{Middle:} We visualize a camera trajectory by plotting camera positions as dots and connecting consecutive frames. 
We show our estimates in \textcolor{cyan}{cyan} and ground truth in \textcolor{green}{green}. 
Our results are highly accurate with very few outliers.
We trained from \rgb images and ground truth poses only, without using the 3D scene model.
 \textbf{Bottom:} Results of competing methods which are less accurate and produce many wrong estimates. 
 To improve the visualization, we connect only consecutive frames within 5 meters range. 
   \textcolor{red}{Red:} In the spirit of PoseNet \cite{kendall2015convolutional}, we train a CNN to predict poses directly. 
   \textcolor{orange}{Orange:} Results of DSAC \cite{brachmann2017dsac} trained with a 3D scene model.
   }

\label{fig:teaser}
\end{figure}

%%%%%%%%% BODY TEXT
\section{Introduction}

Precise localization is key in many applications of computer vision, such as inserting a virtual character into the smartphone view of your living room in a plausible way or guiding a self-driving car through New York. 
While different devices, like smartphones or self-driving cars, come with different senors, an \rgb camera is often present because of its low cost and rich output.
In this work, we present a system which estimates the 6D camera pose, consisting of position and orientation, within a known 3D environment from a single \rgb image.

In the last five years, we have seen an impressive leap forward of what computer vision can achieve due to advances in machine learning.  
Camera localization is a particularly difficult learning problem because training data is usually limited.
Recording a dense or even regular sampling of 6D camera views for any given scene is impractical.
Therefore, generalization to unseen views is a central capability of any camera localization system.

Because of scarce training data, learning the direct mapping of global image appearance to camera pose with a general purpose CNN (convolutional neural net) has seen only limited success. 
Approaches like PoseNet \cite{kendall2015convolutional} and its derivatives \cite{LSTMPoseNet,geometricloss} exhibit low localization accuracy so far. 

An alternative paradigm decomposes camera localization into a sequence of less complex tasks of which only some are learned.
Recently, Brachmann \etal \cite{brachmann2017dsac} presented the differentiable RANSAC (DSAC) pipeline for camera localization.
It builds upon the scene coordinate regression framework originally proposed by Shotton \etal \cite{shotton13scorf}.
The main idea is to map image patches to corresponding points in 3D scene space, so called \emph{scene coordinates}.
This step can be learned even from limited data, since local patch appearance is relatively stable \wrt to view change.
The camera pose, which aligns the image and predicted scene coordinates, can be estimated using RANSAC.

Specifically, in the case of DSAC \cite{brachmann2017dsac}, one CNN predicts scene coordinates, and then random subsets of scene coordinates are used to create a pool of camera pose hypotheses.
Each hypothesis is scored by a second CNN (``scoring CNN'') according to its consensus with the global, \ie image-wide, scene coordinate predictions. 
Based on these scores, one hypothesis is probabilistically chosen, refined and returned as the final camera pose estimate.
The pipeline can be trained end-to-end by optimizing the expected loss of chosen hypotheses.

Brachmann \etal report state-of-the-art accuracy for indoor camera localization \cite{brachmann2017dsac}, but we see three main short-comings of the DSAC pipeline. 
Firstly, the scoring CNN is prone to overfit because it can memorize global patterns of consensus to differentiate good from bad pose hypotheses. 
For example, the CNN might focus on \emph{where} errors occur in the image rather than learning to assess the \emph{quality} of errors.
However, error location does not generalize well to unseen views.
Secondly, initializing the pipeline for \mbox{end-to-end} training requires \rgbd training data or a 3D model of the scene to generate scene coordinate ground truth. 
Neither might be available in some application scenarios.
Thirdly, end-to-end learning is unstable because in the DSAC pipeline \cite{brachmann2017dsac} pose refinement is differentiated via finite differences which leads to high gradient variance.

In this work, we propose a new, fully differentiable camera localization pipeline which has only one learnable component, a fully convolutional neural net for scene coordinate regression.
The output neurons of this network have a limited receptive field, preserving the patch-based nature of scene coordinate prediction.
For hypothesis scoring, we utilize a soft inlier count instead of a learnable CNN.
We show that this simple, differentiable scoring strategy is a reliable measure of pose quality, and yet impervious to overfitting. 
We present a new entropy control method to automatically adapt the magnitude of score values to ensure broad hypotheses distributions for stable end-to-end learning.
We also deploy a new, analytical approximation of pose refinement gradients for additional training stability.
Our pipeline is fast and substantially more accurate than state-of-the-art camera localization methods.

Additionally, and in contrast to previous works \cite{shotton13scorf,valentin2015cvpr, guzman2014multi,cavallari2017fly,brachmann2016,brachmann2017dsac}, we explore learning scene coordinate regression without utilizing a 3D scene model or \rgbd training data. 
\rgbd data might not be available for outdoor scenes, and creating a scene reconstruction often requires tedious trial and error parameter search and manual corrections.
Our system is able to discover an approximate scene geometry automatically due to a coarse initialization followed by optimization of scene coordinate reprojection errors. 
We can still utilize a 3D model if available but do not depend on it for accurate camera localization, see Fig.~\ref{fig:teaser}.

\noindent In the following, we summarize our main contributions.

\begin{compactitem}

\item We present a new camera localization pipeline where a CNN regressing scene coordinates is the only learnable component.  
We implement hypothesis scoring with a new, entropy controlled soft inlier count without learnable parameters, which massively increases generalization capabilities.

\item We show that pose refinement can be effectively differentiated using a local linearization which results in stable end-to-end learning.

\item To the best of our knowledge, we are the first to show that scene coordinate regression can be learned using \rgb images with associated ground truth poses, alone. 
Using a 3D model of the scene is optional since the system can discover scene geometry automatically.

\item We  improve accuracy of \rgb-based 6D camera localization on three datasets, both indoor and outdoor, independent of training with or without a 3D model.
\end{compactitem}

\noindent \textbf{Related Work.} Image-based localization has been addressed using image retrieval techniques, \eg in \cite{schindler2007city} or more recently in \cite{cao2013graph, netvlad2016}.
These methods match a query image to an image database annotated with pose information like geolocation. 
While these methods can scale to extremely large environments, they usually provide only a coarse estimate of the camera location.

Instead of matching to a database, Kendall \etal \cite{kendall2015convolutional} proposed PoseNet, a CNN which learns to map an image directly to a 6D camera pose.
The method has been improved in \cite{geometricloss} by using a reprojection loss, and in \cite{LSTMPoseNet} by using a more expressive architecture.
Although accuracy increased somewhat in these recent works, they are still inferior to competing techniques discussed next.

Accurate 6D camera poses can be recovered using sparse feature-based pipelines. 
Matching local image descriptors to 3D points of a Structure-from-Motion scene reconstruction yields a set of 2D-3D correspondences, from which an accurate pose estimate can be recovered \cite{li2010location}. 
Research focused on making descriptor matching efficient \cite{lim2012real}, robust \cite{svarm2014accurate, sattler2016large} and scale to large outdoor environments \cite{li2016worldwide, sattler2015hyperpoints, sattler2016efficient}.
However, local feature detectors rely on sufficiently textured scenes and good image quality \cite{kendall2015convolutional}. 
Scene reconstruction can also be difficult for some environments \cite{LSTMPoseNet} such as scenes with repeated texture elements.
We surpass the accuracy of state-of-the-art feature-based methods for indoor and outdoor camera localization tasks, often even without using a 3D reconstruction.

The original scene coordinate regression pipeline for \rgbd camera localization of Shotton \etal \cite{shotton13scorf} is related to sparse feature approaches by recovering camera pose by means of 2D-3D correspondences.
But instead of matching a discrete set of points, Shotton \etal formulate correspondence prediction as a continuous regression problem.
A random forest learns to map any image patch to a 3D scene point.  
The scene coordinate regression pipeline has been improved in several follow-up works, \eg in terms of accuracy \cite{valentin2015cvpr, guzman2014multi} or learning camera localization on-the-fly \cite{cavallari2017fly}.
However, these methods heavily depend on a depth channel which greatly simplifies the problem due to strong geometric constrains.
An  \rgb version of the scene coordinate regression pipeline was proposed in \cite{brachmann2016} using an auto-context random forest. 
Similarly, Massiceti \etal \cite{rfvscnn2016} use a random forest mapped to a neural net within a similar pipeline.
However, both systems can only be trained with scene coordinate ground truth, using either \rgbd data or a 3D scene model, and were not trained in an end-to-end fashion.
Our approach is most closely related to the DSAC pipeline \cite{brachmann2017dsac} which was introduced in detail above.

\section{Method}

\begin{figure}[t]
\begin{center}
   \includegraphics[width=1.0\linewidth]{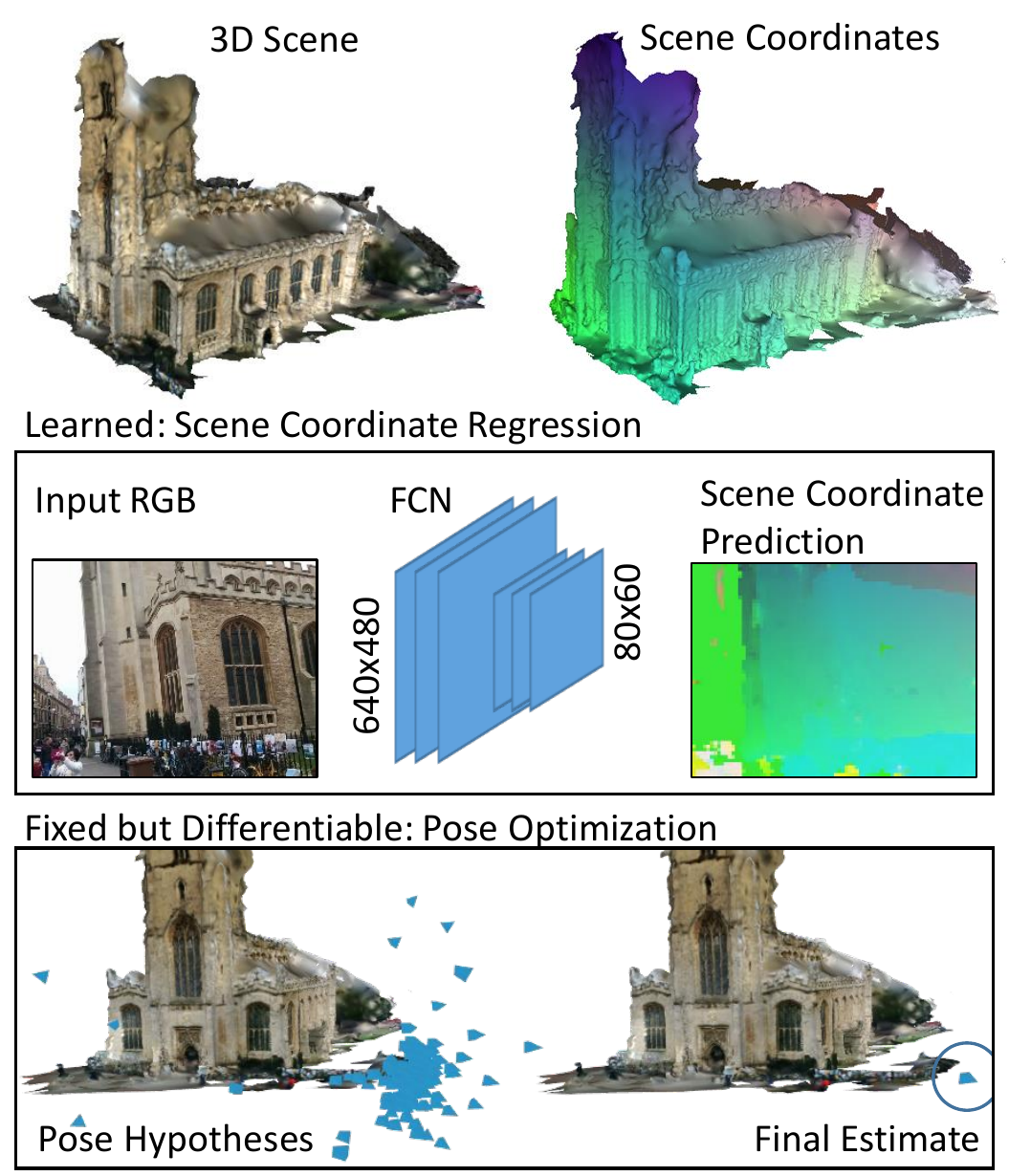}
\end{center}
\vspace{-0.3cm}
   \caption{\textbf{System Overview.} Given an \rgb image, we estimate the 6D camera pose in two stages. Firstly, a fully convolutional network (FCN) regresses 3D scene coordinates (XYZ mapped to \rgb for visualization). This CNN is the only learnable component of our system. Secondly, the system optimizes the pose by sampling a pool of hypotheses, scoring them using a soft inlier count, selecting one according to the scores, and refining it as the final estimate. The second stage contains no learnable parameters but is fully differentiable.}
\label{fig:system}
\end{figure}

Our pipeline follows the basic framework of differentiable RANSAC (DSAC) \cite{brachmann2017dsac} which we describe in Sec.~\ref{sec:meth:bg} for the task of camera pose estimation.
In Sections \ref{sec:meth:reg} and \ref{sec:meth:score}, we explain the key architectural differences to the DSAC pipeline, namely using a fully convolutional network for scene coordinate regression, and scoring pose hypothesis using a soft inlier count without learnable parameters. 
See also Fig.~\ref{fig:system} for an overview of our approach.
We discuss the training procedure of our pipeline, with and without the use of a 3D scene model, in Sec.~\ref{sec:meth:training}.

\subsection{Background}
\label{sec:meth:bg}

Given an \rgb image $I$, we aim at finding an estimate of the 6D camera pose $\opt{\mdl}$ consisting of a 3D translation $\opt{\trans}$ and a 3D rotation $\opt{\rot}$.  
Our system has learnable parameters $\paramw$ which control the search for pose estimate $\opt{\mdl}$.
Differentiable RANSAC \cite{brachmann2017dsac} estimates $\opt{\mdl}$ in the following steps: 

\begin{compactenum}
\item \textbf{Scene Coordinate Regression.} A CNN predicts for each pixel $i$ with position $\pos_i$  the corresponding 3D point $\crd_i(\paramw)$ in the local coordinate frame of the scene. 
This \emph{scene coordinate} $\crd_i(\paramw)$ defines a 2D-3D correspondence between the image and the scene.

\item \textbf{Pose Hypothesis Sampling.} Four scene coordinates suffice to define a unique camera pose by solving the perspective-n-point problem \cite{gao2003complete}.
Since predictions can be erroneous, a pool of $n$ pose hypotheses $\mdl(\paramw)$ is generated by selecting random 4-tupels of scene coordinate predictions. 
Each hypothesis $\mdl(\paramw)$ depends on parameters $\paramw$ via the corresponding scene coordinates.

\item \textbf{Hypothesis Selection.} A function $s(\mdl)$ scores the consensus of each hypothesis with all scene coordinate predictions. 
One hypothesis $\mdl_j(\paramw)$ with index $j$ is selected according to a probability distribution $P(j;\paramw, \alpha)$ which is derived from the score values. 
Hypotheses with a high score are more likely to be selected.
Hyper-parameter $\alpha$ controls the broadness of the distribution, and will be discussed together with the details of scoring in Sec.~\ref{sec:meth:score}.
Selecting hypothesis probabilistically facilitates end-to-end learning, as we will discuss shortly.

\item \textbf{Hypothesis Refinement.} 
Refinement $\refine$ is an iterative procedure which alternates between determining inlier pixels using the current pose estimate, and optimizing the estimate \wrt the inliers.
We discuss the details of refinement in Sec.~\ref{sec:meth:training}.
The selected and refined pose hypothesis is the final estimate of the system, \ie $\opt{\mdl} = \refine(\mdl_j(\paramw))$.
\end{compactenum}

\noindent \textbf{Learning the Pipeline.}
We assume a set of training images $\cal{D}$ with ground truth poses $\gt{\mdl}$.
The probabilistic selection of a pose hypothesis in step 3 allows for optimizing learnable parameters $\paramw$ by minimizing the expected pose loss $\loss$ of the final estimate over the training set \cite{brachmann2017dsac}:
\begin{equation}
\label{eq:dsac}
\opt{\paramw} = \argmin_\paramw \sum_{\cal{D}} \expectation{j \sim P(j;\paramw, \alpha)}{\loss(\refine(\mdl_j(\paramw)), \gt{\mdl})}.
\end{equation}
Any differentiable loss $\loss(\mdl, \gt{\mdl})$ qualifies but we follow \cite{brachmann2017dsac} by using $\loss(\mdl, \gt{\mdl}) = \max(\angle(\rot, \gt{\rot}),||\trans-\gt{\trans}||)$, \ie the maximum of rotational and translational error.
Partial derivatives of Eq.~\ref{eq:dsac} \wrt parameters $\paramw$ are given by 
\begin{equation}
\label{eq:ddsac}
\derv{\paramw} \expectation{j}{\cdot} = \expectation{j}{\loss(\cdot) \derv{\paramw} \log P(j;\paramw, \alpha) + \derv{\paramw}\loss(\cdot)},
\end{equation}
where we use ($\cdot$) as a stand-in for corresponding arguments of Eq.~\ref{eq:dsac}.
Eq.~\ref{eq:ddsac} allows us to learn our pipeline in an \mbox{end-to-end} fashion using standard back propagation.
\subsection{Scene Coordinate Regression}
\label{sec:meth:reg}

\begin{figure}[th!]
\begin{center}
   \includegraphics[width=0.9\linewidth]{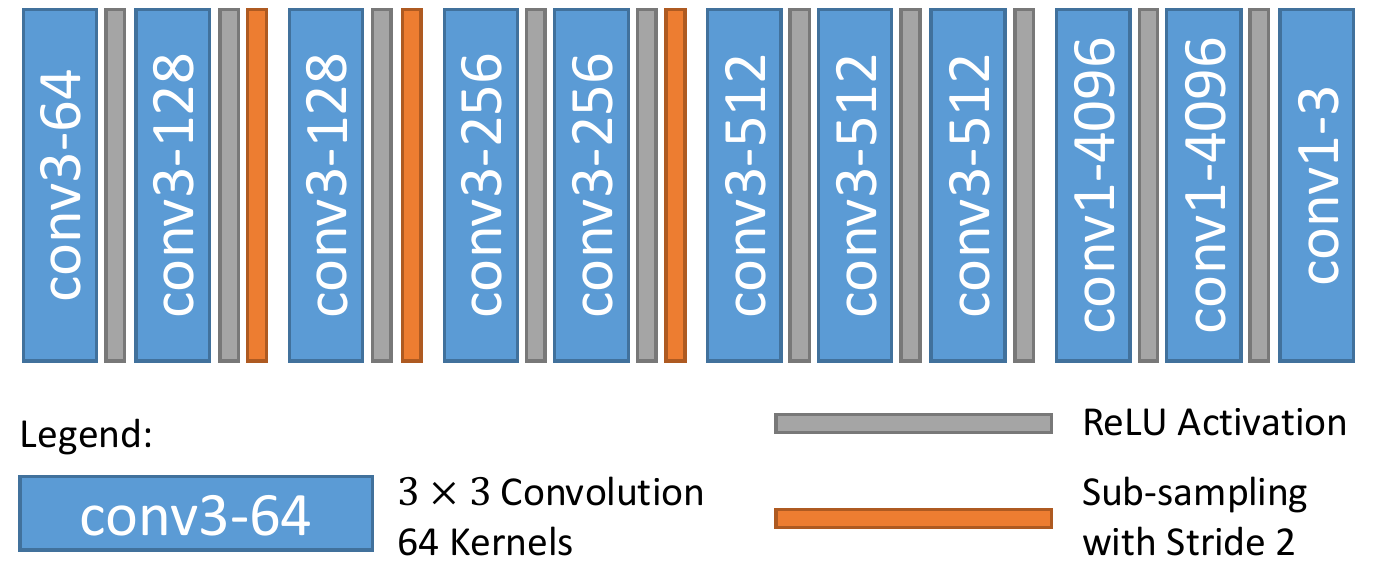}
\end{center}

   \caption{\textbf{Network Architecture.} Our FCN network architecture is composed of $3 \times 3$ convolutions, ReLU activations and sub-sampling via strided convolutions. The final fully connected layers are implemented via $1 \times 1$ convolutions.}
\label{fig:architecture}
\end{figure}

The DSAC pipeline \cite{brachmann2017dsac} uses a CNN for scene coordinate regression which takes an image patch of $42 \times 42$ px as input and produces one scene coordinate prediction for the center pixel. 
This design is inefficient because the CNN processes neighboring patches independently without reusing computations. 
They alleviate the problem by sampling $40 \times 40$ patches per image instead of making a dense prediction for all patches.
In contrast, we use a fully convolutional network (FCN) \cite{fcn2015}, although without upsampling layers.
Our FCN takes an \rgb image of $640 \times 480$ px as input and produces $80 \times 60$ scene coordinate predictions, \ie we regress more scene coordinates in less time. 
Similar to DSAC \cite{brachmann2017dsac}, we use a VGG-style \cite{Simonyan2014} architecture with $\approx$ 30M parameters, and our output neurons have a receptive field of $41 \times 41$ px.
See Fig.~\ref{fig:architecture} for a schematic of the FCN architecture.

\subsection{Hypothesis Scoring}
\label{sec:meth:score}

Scoring determines which camera pose hypothesis is chosen and refined to yield the final estimate.
DSAC \cite{brachmann2017dsac} uses a separate CNN for this task. 
This scoring CNN takes a $40 \times 40$ image of reprojection errors, and regresses a score value $s(\mdl)$ for each hypothesis.
The reprojection error for pixel $i$ and hypothesis $\mdl$ is defined as 
\begin{equation}
\label{eq:inlier}
r_i(\mdl, \paramw) = ||C\mdl^{-1}\crd_i(\paramw) - \pos_i||,
\end{equation} 
where $C$ is the camera calibration matrix.
We assume homogeneous coordinates, and application of perspective division before calculating the norm.
The final hypothesis is chosen according to the softmax distribution $P(j;\paramw, \alpha)$:
\begin{equation}
j\sim P(j;\paramw, \alpha) = \frac{\exp(\alpha s(\mdl_j(\paramw)))}{\sum_k \exp(\alpha s(\mdl_k(\paramw)))},
\end{equation} 
where hyper-parameter $\alpha$ is a fixed scaling factor that ensures a broad distribution.
Parameter $\alpha$ controls the flow of gradients in end-to-end learning by limiting or enhancing the influence of hypotheses with smaller scores compared to hypotheses with larger scores.

\begin{figure*}[t]
\begin{center}
   \includegraphics[width=0.9\linewidth]{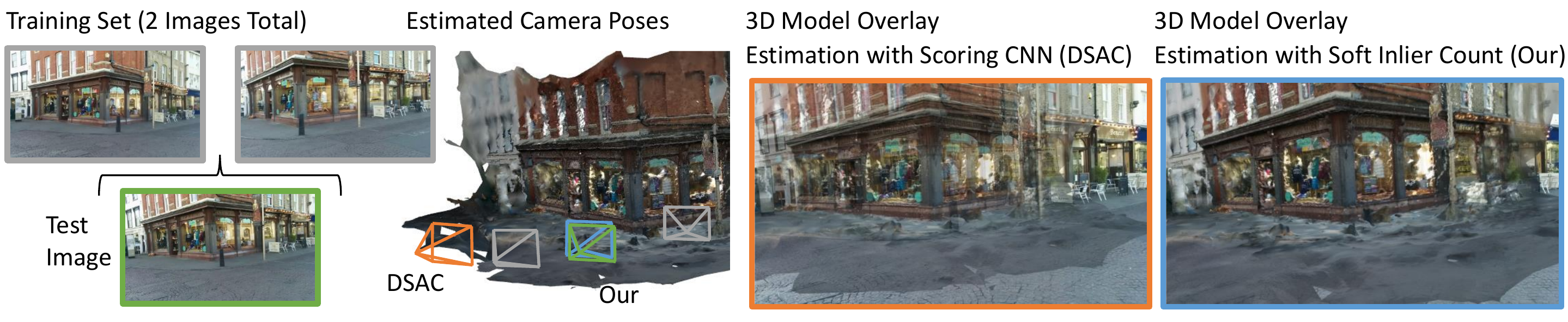}
\end{center}
   \caption{\textbf{Generalization Capabilities.} We train two CNNs for scene coordinate regression and pose hypothesis scoring following DSAC~\cite{brachmann2017dsac} but using only two training images (\textcolor{gray}{gray}). The system cannot generalize to an unseen test image (\textcolor{green}{green}), and produces an estimate far off (\textcolor{orange}{orange}, 3.7m and 12.6$^\circ$ pose error). Exchanging the scoring CNN for a soft inlier count (see Sec.~\ref{sec:meth:score}), we obtain an accurate estimate (\textcolor{blue}{blue}, 0.1m and 0.3$^\circ$ pose error). Note that, apart from the scoring function, we fix all components of the test run, including pose hypotheses sampling. This experiment illustrates that scene coordinate regression generalizes well, while score regression does not.}
\label{fig:overfit}
\end{figure*}

We identify two problems when learning function $s(\mdl)$.
Firstly, the image of reprojection errors contains information about the global image structure, \eg \emph{where} errors occur. 
Scene coordinate regression generalizes well because only local, \ie patch-based, predictions are being made.
The scoring CNN of DSAC \cite{brachmann2017dsac} however learns patterns in the global error image that do not generalize well to unseen views.
Secondly, during end-to-end learning, the score CNN has an incentive to produce increasingly large scores which puts more weight on the best hypothesis over all other hypotheses in the pool.
At some point, one hypothesis will have probability 1 resulting in the minimal expected task loss, see Eq.~\ref{eq:dsac}. 
Note that the scaling factor $\alpha$ is fixed in \cite{brachmann2017dsac}.
Hence, distribution $P(j;\paramw, \alpha)$ can collapse, leading to overfitting and training instability.
Regularization might be able to alleviate both problems but we show that a simple and differentiable inlier counting schema is an effective and robust measure of camera pose quality, see Fig.~\ref{fig:overfit}.

\noindent \textbf{Soft Inlier Counting.}
The original RANSAC-schema \cite{ransac1981} measures hypothesis consensus by counting data points explained by the model, so called inliers.
In our application, the inlier count is given by $\sum_i  \mathbbm{1}(\tau - r_i(\mdl,\paramw))$, where $\mathbbm{1}$ denotes the Heaviside step function, $r_i$ denotes a reprojection error (see Eq.~\ref{eq:inlier}), and $\tau$ is the inlier threshold.
Several earlier scene coordinate regression works relied on the inlier count to score hypotheses, \eg \cite{shotton13scorf,brachmann2016}.
For our purpose, we construct a differentiable version by substituting the step function with a sigmoid function.
\begin{equation}
\label{eq:score}
s(\mdl) = \sum_i \text{sig}( \tau - \beta (r_i(\mdl,\paramw))),
\end{equation}
where hyper-parameter $\beta$ controls the softness of the sigmoid. 
This scoring function is similar to MSAC \cite{mlesac2000} but does not use a hard cut-off.
We use Eq.~\ref{eq:score} in distribution $P(j;\paramw, \alpha)$ to select a pose hypothesis.

\noindent \textbf{Controlling Entropy.}
The magnitude of inlier scores can vary significantly depending on the difficulty of the scene, usually ranging from $10^2$ to $10^3$ for different environments.
The magnitude can also change during end-to-end learning when scene coordinate regression improves and produces smaller reprojection errors.
As mentioned earlier, keeping scores within a reasonable range is important for having a broad distribution $P(j;\paramw, \alpha)$, and hence stabilizing \mbox{end-to-end} training.
Setting $\alpha$ manually per scene is a tedious task, hence we adapt $\alpha$ automatically during \mbox{end-to-end} training. 
We measure distribution broadness via the Shannon entropy as a function of $\alpha$: 
\begin{equation}
S(\alpha) = - \sum_j P(j;\paramw,\alpha) \log P(j;\paramw,\alpha).
\end{equation}
We optimize $\alpha$ according to $\argmin_\alpha |S(\alpha) - \gt{S}|$ with target entropy value $\gt{S}$ via gradient descent in parallel to \mbox{end-to-end} training of the pipeline. 
This schema establishes the target entropy within the first few iterations of \mbox{end-to-end} training and keeps it stable throughout.

\subsection{Training Procedure}
\label{sec:meth:training}

Our pipeline can be trained in an end-to-end fashion using pairs of \rgb images and ground truth poses, but doing so from scratch will fail as the system quickly reaches a local minimum.
The DSAC pipeline is initialized using scene coordinate ground truth extracted from \rgbd training data \cite{brachmann2017dsac}.
We propose a new 3-step training schema with different objective functions in each step.
Depending on whether a 3D scene model is available or not, we use rendered or approximate scene coordinates to initialize the network in the first step.
Training steps two and three improve the accuracy of the system which is crucial when no 3D model was provided for initialization.

\begin{figure}[t!]
\begin{center}
   \includegraphics[width=0.9\linewidth]{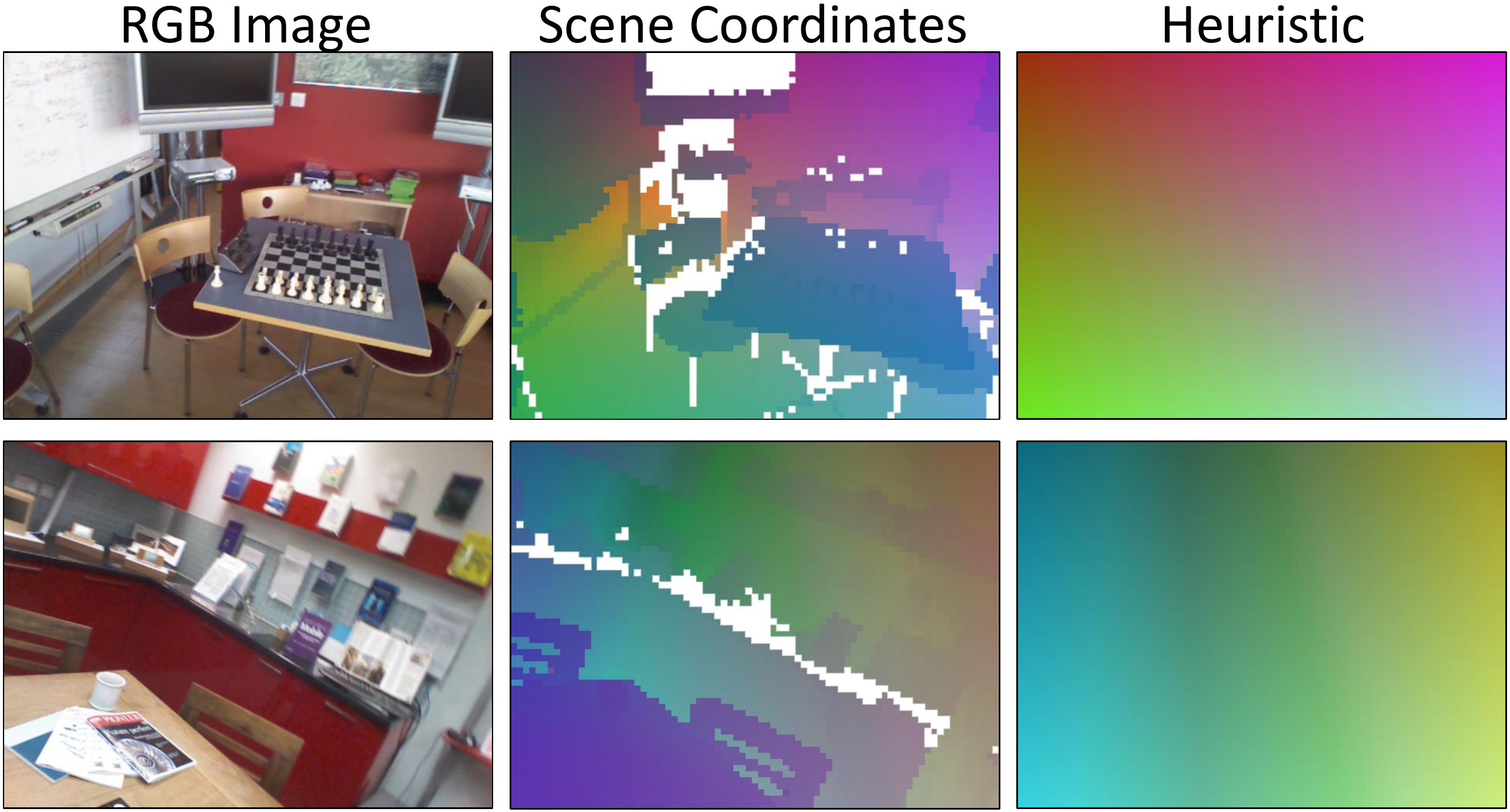}
\end{center}
   \caption{\textbf{Initial Scene Coordinate Heuristic}. When ground truth scene coordinates (middle) are unknown, we use a heuristic (right) for initializing our pipeline. 
   This heuristic assumes that each scene point has a constant distance to the camera plane.}
\label{fig:heuristic}
\end{figure}

\noindent \textbf{Scene Coordinate Initialization.} In the first training step, we initialize our pipeline similar to DSAC \cite{brachmann2017dsac} by optimizing
\begin{equation}
\opt{\paramw} = \argmin_\paramw \sum_{i} ||\crd_i(\paramw) - \gt{\crd}_i||.
\end{equation}
We render scene coordinates $\gt{\crd}$ using ground truth poses $\gt{\mdl}$, and a 3D scene model, if available.
Without a 3D model, we approximate scene coordinate $\gt{\crd}_i$ by $\gt{\mdl} (\frac{dx_i}{f}, \frac{dy_i}{f},d,1)^T$, where $x_i$ and $y_i$ are the 2D coordinates of pixel $i$, $f$ denotes the focal length, and $d$ represents a constant depth prior. 
This heuristic assumes that all scene points have a constant distance from the camera plane,
see Fig.~\ref{fig:heuristic} for a visualization.
The heuristic ignores scene geometry completely.
However, it effectively disentangles camera views by coarsely assigning the correct range of scene coordinate values to different spatial parts of the 3D environment.
The heuristic itself will yield poor localization accuracy, but serves as basis for the next training step.

\noindent \textbf{Optimization of Reprojection Error.} In a second training step, we optimize the reprojection error, calculated using the ground truth pose. 
It encodes single view constraints that are effective for recovering the correct depth of scene points in case we used the heuristic in the first training step.
Thus, we optimize
\begin{equation}
\opt{\paramw} = \argmin_\paramw \sum_{i} r_i(\gt{\mdl}, \paramw),
\end{equation}
which lets our system learn about the scene geometry without using a 3D scene model.
However, we found that optimizing the reprojection error can improve accuracy, even if a 3D model was available for the initialization.

\noindent \textbf{End-to-End Optimization.}
In a third step, we train our pipeline in an end-to-end fashion according to Eq.~\ref{eq:dsac}.
This requires all components to be differentiable, including pose refinement.
In DSAC \cite{brachmann2017dsac}, refinement gradients are calculated via finite differences.
Note that refinement depends on thousands of inputs, namely all scene coordinate predictions, which makes calculation of finite differences slow, and results in high gradient variance due to numerical instabilities.
In the following, we discuss pose refinement in detail, and explain an efficient, analytical approximation for refinement gradients, resulting in stable end-to-end training.

As mentioned earlier, pose refinement alternates between, firstly, determining a set of inlier pixels \wrt to the current pose estimate, and, secondly, optimizing the pose \wrt to reprojection errors over the inlier set.
We define the inlier set ${\cal{I}}$ to contain all pixels with a reprojection error $r_i$ below a threshold $\tau$, \ie ${\cal{I}} = \{ i| r_i < \tau \}$.
For pose optimization, we combine the reprojection errors of all inliers within one residual vector: 
\begin{equation}
(\mathbf{r}(\mdl, \paramw))_i =
\begin{cases} 
      r_i(\mdl, \paramw) & \text{if $i \in \cal{I}$}, \\
      0 & \text{otherwise}
   \end{cases}
\end{equation}
We optimize the pose according to: 
\begin{equation}
\refine(\mdl) = \argmin_{\mdl'} ||\mathbf{r}(\mdl', \paramw)||^2.
\end{equation}
We use the iterative Gauss-Newton algorithm, which gives the following update rule:
\begin{equation}
\refine^{t+1} = \refine^t - (J_\res^T J_\res)^{-1}J_\res^T \res(\refine^t, \paramw),
\end{equation}
where $t$ denotes the iteration number, and we abbreviate $\refine^t(\mdl)$ by $\refine^t$.
The Jacobean matrix $J_\res$ contains partial derivatives $(J_\res)_{ij} = \frac{\partial (\res(\refine^t, \paramw))_i}{\partial (\refine^t)_j}$.
We optimize until convergence, re-calculate the inlier set, and repeat until the inlier set converges, too.
Note that for DSAC \cite{brachmann2017dsac} the number of refinement iterations and the number of inliers considered are limited to reduce the computational demand of finite differences.
We do not have to make similar concessions, here.

The model linearization of Gauss-Netwon allows us to approximate refinement gradients $\derv{\paramw}\refine(\mdl)$ around the optimum found in the last optimization iteration \cite{Foerstner2016Photogrammetric}. 
We fix the optimum as $\mdl_\text{O} = \refine^{t=\infty}(\mdl)$ which makes the last update step
$\refine(\mdl) = \mdl_\text{O} - (J_\res^T J_\res)^{-1}J_\res^T \res(\mdl_\text{O}, \paramw)$,
and the corresponding derivatives 
\begin{equation}
\derv{\paramw}\refine(\mdl) \approx - (J_\res^T J_\res)^{-1}J_\res^T \derv{\paramw}\res(\mdl_\text{O}, \paramw),
\end{equation}
which allow for stable end-to-end training according to Eq.~\ref{eq:dsac}.

\section{Experiments}

\begin{table*}[t!h]
%\vspace{-0.5cm}
\centering
\caption{\textbf{Median 6D Localization Errors.} We report results for the 7Scenes dataset \cite{shotton13scorf} and the Cambridge Landmarks dataset \cite{kendall2015convolutional}. We mark best results \textbf{bold} (if both, translational and rotational error, are lowest). Results of DSAC marked with an asterisk (*) are before end-to-end optimization which did not converge. A dash (-) indicates that a method failed completely.}
\label{res:main}
\begin{tabular}{@{}l|cccc|ccc@{}}
Dataset                 & \multicolumn{4}{l|}{Training w/ 3D Model}              & \multicolumn{1}{l}{w/o 3D Model} & \multicolumn{1}{l}{}             & \multicolumn{1}{l}{} \\ \midrule
7Scenes            & \begin{tabular}[c]{@{}c@{}}PoseNet \cite{geometricloss} \\ \small{(Geom.~Loss)}\end{tabular}    & \begin{tabular}[c]{@{}c@{}}Active \\ Search \cite{sattler2016efficient}\end{tabular} & \begin{tabular}[c]{@{}c@{}}DSAC \cite{brachmann2017dsac} \\ \small{(\rgb Training)}  \end{tabular}      & Ours        & \begin{tabular}[c]{@{}c@{}}PoseNet \cite{geometricloss} \\ \small{(Pose Loss)}\end{tabular}                         & \begin{tabular}[c]{@{}c@{}}Spatial \\ LSTM \cite{LSTMPoseNet}\end{tabular} & Ours                  \\ \midrule
Chess   & 0.13m, 4.5$^\circ$  & 0.04m, 2.0$^\circ$      & 0.02m, 1.2$^\circ$    & \textbf{0.02m, 0.5$^\circ$}  & 0.14m, 4.5$^\circ$      & 0.24m, 5.8$^\circ$ & \textbf{0.02m, 0.7$^\circ$}        \\
Fire    & 0.27m, 11.3$^\circ$ & 0.03m, 1.5$^\circ$      & 0.04m, 1.5$^\circ$    & \textbf{0.02m, 0.9$^\circ$}  & 0.27m, 11.8$^\circ$     & 0.34m, 11.9$^\circ$ & \textbf{0.03m, 1.1$^\circ$}        \\
Heads   & 0.17m, 13.0$^\circ$ & 0.02m, 1.5$^\circ$      & 0.03m, 2.7$^\circ$    & \textbf{0.01m, 0.8$^\circ$}  & 0.18m, 12.1$^\circ$     & 0.21m, 13.7$^\circ$ & \textbf{0.12m, 6.7$^\circ$}      \\
Office  & 0.19m, 5.6$^\circ$  & 0.09m, 3.6$^\circ$      & 0.04m, 1.6$^\circ$    & \textbf{0.03m, 0.7$^\circ$}  & 0.20m, 5.8$^\circ$      & 0.30m, 8.1$^\circ$ & \textbf{0.03m, 0.8$^\circ$}        \\
Pumpkin & 0.26m, 4.8$^\circ$  & 0.08m, 3.1$^\circ$      & 0.05m, 2.0$^\circ$    & \textbf{0.04m, 1.1$^\circ$}  & 0.25m, 4.8$^\circ$      & 0.33m, 7.0$^\circ$ & \textbf{0.05m, 1.1$^\circ$}        \\
Kitchen & 0.23m, 5.4$^\circ$  & 0.07m, 3.4$^\circ$      & 0.05m, 2.0$^\circ$    & \textbf{0.04m, 1.1$^\circ$}  & 0.24m, 5.5$^\circ$      & 0.37m, 8.8$^\circ$ & \textbf{0.05m, 1.3$^\circ$}        \\
Stairs  & 0.35m, 12.4$^\circ$ & \textbf{0.03m, 2.2$^\circ$}      & 1.17m, 33.1$^\circ$ & 0.09m, 2.6$^\circ$ & 0.37m, 10.6$^\circ$     & 0.40m, 13.7$^\circ$ & \textbf{0.29m, 5.1$^\circ$}  \\ \midrule
Cambridge & \multicolumn{7}{l}{ } \\ \midrule
Great Court      & 7.00m, 3.7$^\circ$  & -             & *2.80m, 1.5$^\circ$ & \textbf{0.40m, 0.2$^\circ$} & 6.83m, 3.5$^\circ$                       & -                                & \textbf{0.66m, 0.4}$^\circ$          \\
K.~College   & 0.99m, 1.1$^\circ$  & 0.42m, 0.6$^\circ$    & *0.30m, 0.5$^\circ$ & \textbf{0.18m, 0.3$^\circ$} & 0.88m, 1.0$^\circ$                       & 0.99m, 1.0$^\circ$                       & \textbf{0.23m, 0.4$^\circ$}           \\
Old Hospital     & 2.17m, 2.9$^\circ$  & 0.44m, 1.0$^\circ$    & 0.33m, 0.6$^\circ$  & \textbf{0.20m, 0.3$^\circ$} & 3.20m, 3.3$^\circ$                       & 1.51m, 4.3$^\circ$                       & \textbf{0.24m, 0.5$^\circ$}         \\
Shop Facade      & 1.05m, 4.0$^\circ$  & 0.12m, 0.4$^\circ$    & 0.09m, 0.4$^\circ$  & \textbf{0.06m, 0.3$^\circ$} & 0.88m, 3.8$^\circ$                       & 1.18m, 7.4$^\circ$                       & \textbf{0.09m, 0.4$^\circ$}           \\
St M.~Church & 1.49m, 3.4$^\circ$  & 0.19m, 0.5$^\circ$    & *0.55m, 1.6$^\circ$ & \textbf{0.13m, 0.4$^\circ$} & 1.57m, 3.2$^\circ$                       & 1.52m, 6.7$^\circ$                       & \textbf{0.20m, 0.7$^\circ$}           \\
Street           & 20.7m, 25.7$^\circ$ & \textbf{0.85m, 0.8$^\circ$}    & -           & -          & \textbf{20.3m, 25.5$^\circ$}                      & -                                & -                   
\end{tabular}
\end{table*}

\begin{figure*}[h!]
\begin{center}
   \includegraphics[width=1.0\linewidth]{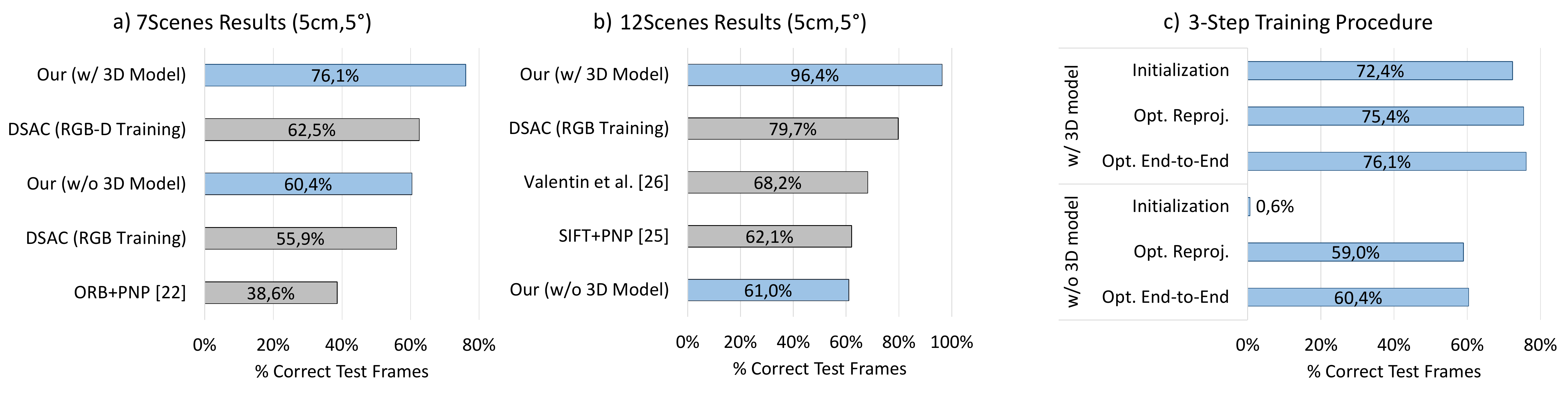}
\end{center}
   \caption{\textbf{Indoor Localization Accuracy.} We show the percentage of test frames of the 7Scenes \textbf{(a)} \resp the 12Scenes \textbf{(b)} dataset with a pose error below 5cm and 5$^\circ$. We mark our method blue. Note that all competitors depend on a 3D model.
For our method, we show results after each of our 3 training steps on the 7Scenes dataset \textbf{(c)}.}
\label{fig:5cm5deg}
%\vspace{-0.5cm}
\end{figure*}

We evaluate our approach on three publicly available camera localization datasets, both indoor and outdoor. 

\noindent \textbf{7Scenes \cite{shotton13scorf}.} 
This \rgbd dataset comprises of 7 difficult indoor scenes with motion blur, repeating structures and texture-less surfaces.
Several thousand frames are given as training and test sequences for each scene. The dataset includes ground truth poses and accurate 3D scene models.
We ignore depth channels and utilize only \rgb images. 

\noindent \textbf{12Scenes \cite{valentin2016learning}.} 
This dataset is very similar to 7Scenes but has larger indoor environments, and contains smaller training sets of several hundred frames per scene.

\noindent \textbf{Cambridge Landmarks \cite{kendall2015convolutional}.} 
The dataset contains \rgb images of six large outdoor environments, divided in training and test sequences of several hundred frames.
Coarse SfM reconstructions are also provided. 

\noindent \textbf{Parameter Settings.} 
We train our pipeline for a fixed number of iterations using ADAM~\cite{adam2014} on full training sets, and select hyper-parameters that achieve the lowest training loss.
During test time we always choose the hypothesis with maximum score.
See appendix \ref{sec:parameters} for a full parameter listing.

\noindent \textbf{Competitors.}
We compare to the latest incarnation of PoseNet \cite{geometricloss} which can be trained using a standard pose loss or, utilizing a 3D model, a geometric loss for improved accuracy.
We also compare to the Spatial LSTM of \cite{LSTMPoseNet}.
We include results of several sparse feature baselines, most notably Active Search \cite{sattler2016efficient}.
We compare to DSAC~\cite{brachmann2017dsac} trained using \rgbd training data as in \cite{brachmann2017dsac}, and using rendered scene coordinates (denoted ``RGB training'').

\subsection{Camera Localization Accuracy}

We list our main experimental results in Table~\ref{res:main} for the 7Scenes and Cambridge datasets.
Compared to the PoseNet variants \cite{geometricloss,LSTMPoseNet} we improve accuracy by a factor of 10 for many scenes, and compared to the sparse feature-based Active Search \cite{sattler2016efficient} by a factor of 2.
Compared to DSAC \cite{brachmann2017dsac}, which is the strongest competitor, we massively improve accuracy for the Cambridge dataset. 
We observe only a small to moderate loss in accuracy when our method is trained without a 3D scene model.
Note that the only two competitors that do not depend on a 3D model, namely PoseNet \cite{geometricloss} and the spatial LSTM \cite{LSTMPoseNet}, achieve a much lower accuracy.
In fact, for most scenes, our method trained without a 3D model surpasses the accuracy of competitors utilizing a 3D model.
Similar to DSAC \cite{brachmann2017dsac} and the Spatial LSTM \cite{LSTMPoseNet}, we were not able to estimate reasonable poses for the Cambridge Street scene which is one order of magnitude larger than the other outdoor scenes.
The capacity of our neural network might be insufficient for this scene scale but we did not explore this possibility.

The median pose accuracy used in Table \ref{res:main} does not reflect the frequency of wrong pose estimates very well. 
Therefore, we show the percentage of test images with a pose error below 5cm and 5$^\circ$ for the 7Scenes dataset in Fig.~\ref{fig:5cm5deg} a).
Note that all competitors listed require a 3D model of the scene.
PoseNet \cite{geometricloss} and the Spatial LSTM \cite{LSTMPoseNet} do not report results using this measure, but based on their median accuracy they achieve less than 50\% on 7Scenes. 
We outperform all competitors, most notably DSAC \cite{brachmann2017dsac} trained with \rgbd data (+13.6\%).
When training DSAC using a 3D model (``RGB Training''), its performance drops by 6.6\% due to inaccuracies in the 3D model.
Our method trained without a 3D model exceeds the accuracy of DSAC trained with a 3D model by 4.5\%.

We show results for the 12Scenes dataset in \mbox{Fig.~\ref{fig:5cm5deg} b)}.
We achieve an accuracy of 96.4\% with a 16.7\% margin to DSAC.
Training without a 3D model still achieves a good accuracy comparable to a sparse feature baseline (SIFT+PNP \cite{valentin2016learning}).

\begin{figure*}[h!]
\begin{center}
\vspace{0.5cm}
   \includegraphics[width=0.9\linewidth]{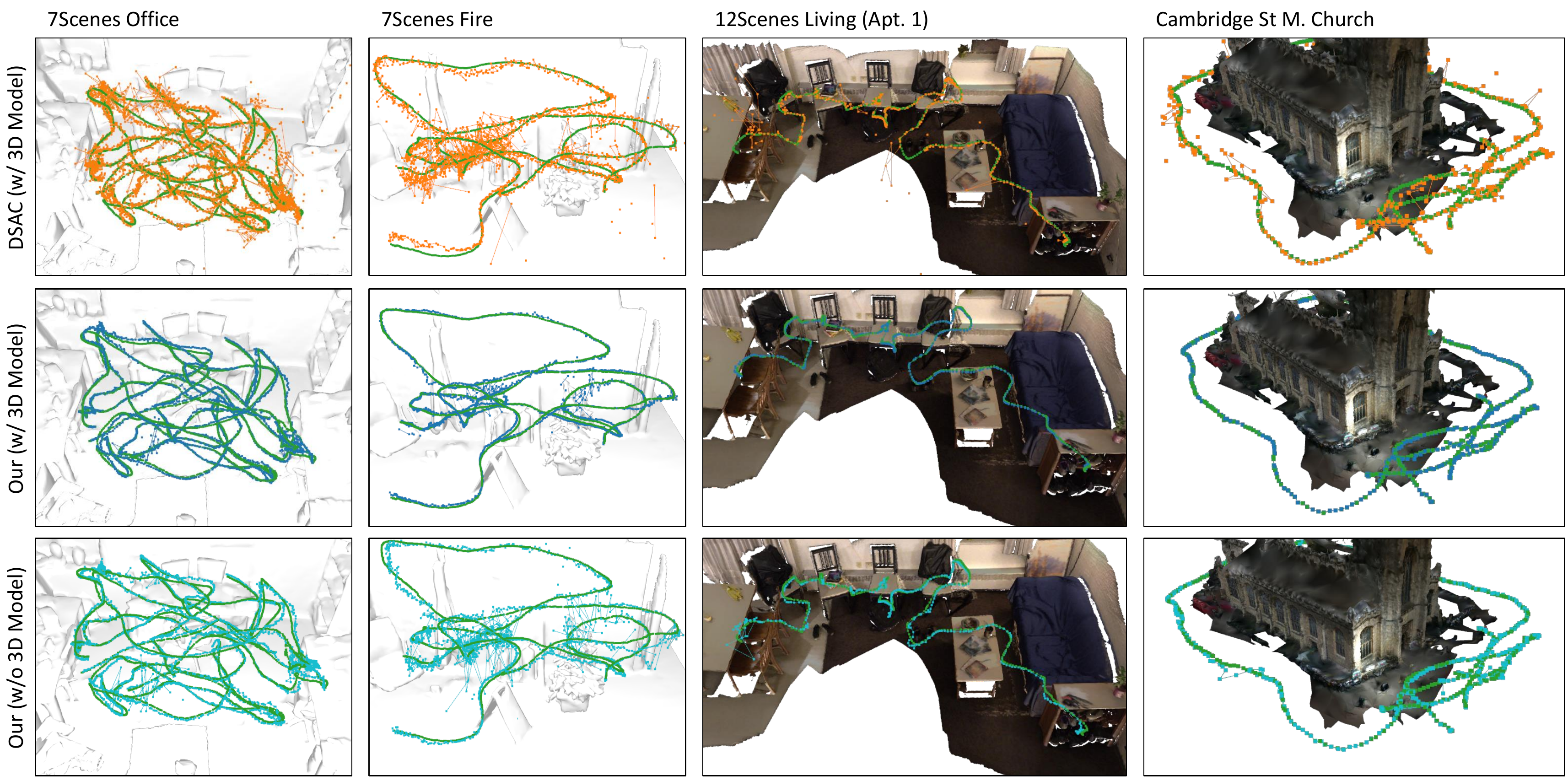}
\end{center}
   \caption{\textbf{Estimated Camera Trajectories}. We plot estimated camera locations as trajectories within the respective 3D scene model (untextured for 7Scenes). 
   We show ground truth in \textcolor{green}{green}, DSAC \cite{brachmann2017dsac} estimates in \textcolor{orange}{orange}, and our results in \textcolor{blue}{blue} and \textcolor{cyan}{cyan} when trained with and without a 3D model, respectively. Note that DSAC produces many wrong estimates despite being trained with a 3D model.}
\label{fig:quality}
%\vspace{-0.5cm}
\end{figure*}

See Fig.~\ref{fig:quality} for a qualitative comparison of DSAC and our method trained with and without a 3D model, respectively. 
We include additional qualitative results in the supplementary video\footnote{\url{https://youtu.be/DjJFRTFEUq0}}. 
See Table \ref{table:7scenesdetails} (left) for the accuracy of our pipeline for each scene of the 7Scenes dataset \cite{shotton13scorf}, when trained with and without a 3D scene model.
Table \ref{table:12scenesdetails} shows the corresponding results for the 12Scenes dataset \cite{valentin2016learning}.

\begin{table*}[h!t!]
\centering
\caption{\textbf{Detailed Results for 7Scenes.} We report accuracy per scene as percentage of estimated poses with an error below 5cm and 5$^\circ$. ``Complete'' denotes the accuracy on all test frames of the dataset combined. \textbf{Left:} Results of training the pipeline with and without a 3D model. Numbers are given after end-to-end training. \textbf{Right:} Effect of varying the output resolution of scene coordinate regression (default: 80x60). Numbers are given for training with a 3D model and after end-to-end training.}
\label{table:7scenesdetails}
\begin{tabular}{l|cc||cccccc|}
\cline{2-9}
                               & \multicolumn{2}{c||}{Training} & \multicolumn{6}{c|}{Scene Coordinate Output Resolution} \\  
                               & w/ 3D Model   & w/o 3D Model  & 20x15   & 20x30   & 40x30   & 40x60  & 80x60  & 320x240 \\ \hline
\multicolumn{1}{|l|}{Chess}    & 97.1\%        & 93.8\%        & 96.5\%  & 97.1\%  & 97.3\%  & 97.4\% & 97.1\% & 97.6\%  \\
\multicolumn{1}{|l|}{Fire}     & 89.6\%        & 75.6\%        & 86.0\%  & 87.8\%  & 89.5\%  & 90.4\% & 89.6\% & 91.9\%  \\
\multicolumn{1}{|l|}{Heads}    & 92.4\%        & 18.4\%        & 88.6\%  & 89.9\%  & 91.3\%  & 92.1\% & 92.4\% & 93.7\%  \\
\multicolumn{1}{|l|}{Office}   & 86.6\%        & 75.4\%        & 83.3\%  & 85.2\%  & 86.2\%  & 86.1\% & 86.6\% & 87.3\%  \\
\multicolumn{1}{|l|}{Pumpkin}  & 59.0\%        & 55.9\%        & 58.2\%  & 60.5\%  & 61.3\%  & 61.3\% & 59.0\% & 61.6\%  \\
\multicolumn{1}{|l|}{Kitchen}  & 66.6\%        & 50.7\%        & 63.2\%  & 64.3\%  & 64.7\%  & 64.7\% & 66.6\% & 65.7\%  \\
\multicolumn{1}{|l|}{Stairs}   & 29.3\%        & 2.0\%         & 19.7\%  & 22.5\%  & 23.7\%  & 25.1\% & 29.3\% & 28.7\%  \\ \hline
\multicolumn{1}{|l|}{Complete} & \textbf{76.1\%}        & \textbf{60.4\%}        & \textbf{72.8\%}  & \textbf{74.4\%}  & \textbf{75.2\%}  & \textbf{75.5\%} & \textbf{76.1\%} & \textbf{76.6\%}  \\ \hline
\end{tabular}
%\vspace{-0.5cm}
\end{table*}

\begin{table}[t!]
\centering
\caption{\textbf{Detailed Results for 12Scenes.} We report accuracy per scene as percentage of estimated poses with an error below 5cm and 5$^\circ$. ``Complete'' denotes the accuracy on all test frames of the dataset combined. Numbers are given for training the pipeline with and without a 3D model, both after end-to-end training.}
\label{table:12scenesdetails}
\begin{tabular}{ll|cc|}
\cline{3-4}
                                                &           & \multicolumn{2}{c|}{Training} \\
                                                &           & w/ 3D Model   & w/o 3D Model  \\ \hline
\multicolumn{1}{|l|}{\multirow{2}{*}{Apt.~1}}   & Kitchen   & 100\%         & 7.6\%         \\
\multicolumn{1}{|l|}{}                          & Living    & 100\%         & 92.0\%        \\ \hline
\multicolumn{1}{|l|}{\multirow{4}{*}{Apt.~2}}   & Bed       & 99.5\%        & 66.1\%        \\
\multicolumn{1}{|l|}{}                          & Kitchen   & 99.5\%        & 87.6\%        \\
\multicolumn{1}{|l|}{}                          & Living    & 100\%         & 89.9\%        \\
\multicolumn{1}{|l|}{}                          & Luke      & 95.5\%        & 67.3\%        \\ \hline
\multicolumn{1}{|l|}{\multirow{4}{*}{Office 1}} & Gates 362 & 100\%         & 96.3\%        \\
\multicolumn{1}{|l|}{}                          & Gates 381 & 96.8\%        & 27.8\%        \\
\multicolumn{1}{|l|}{}                          & Lounge    & 95.1\%        & 94.8\%        \\
\multicolumn{1}{|l|}{}                          & Manolis   & 96.4\%        & 72.2\%        \\ \hline
\multicolumn{1}{|l|}{\multirow{2}{*}{Office 2}} & Floor 5a  & 83.7\%        & 11.0\%        \\
\multicolumn{1}{|l|}{}                          & Floor 5b  & 95.0\%        & 83.2\%        \\ \hline
\multicolumn{2}{|c|}{Complete}                              & \textbf{96.4\%}        & \textbf{60.9\%}        \\ \hline
\end{tabular}
\end{table}

\subsection{Detailed Studies}

\noindent \textbf{Inlier Count vs.~Scoring CNN.}
We retrain DSAC, substituting the scoring CNN for our soft inlier count. 
We measure the percentage of test frames with an error below 5cm and 5$^\circ$.
Results improve from 55.9\% to 58.9\% for 7Scenes. 
The effect is strongest for the \emph{Heads} and \emph{Stairs} scenes (+19\% \resp +8\%) which have the smallest training sets.
Accuracy for 12Scenes, where all training sets are small, also increases from 79.7\% to 89.6\%.
We conclude that the soft inlier count helps generalization, considerably.
The scoring CNN easily overfits to the spatial constellation of reprojection errors. 
For large training sets (\eg 7Scenes Kitchen) accuracy differs by less than 1\% between the two scoring methods, but inlier counting is always superior. 
We expect the accuracy gap to vanish completely given enough training data, but at the moment we see no evidence that the scoring CNN could learn to make a more intelligent decision than the inlier counting schema.

\noindent \textbf{Impact of Training Steps.}
See Fig.~\ref{fig:5cm5deg} c) for a breakdown of our 3-step training procedure. 
When a 3D scene model is available for training, the first training step already achieves high accuracy.
Optimizing the reprojection error and \mbox{end-to-end} training improve accuracy further.
When no 3D model is available, optimizing the reprojection error is imperative for good results, since  discovering scene geometry is necessary to generalize to unseen views. 
End-to-end training can additionally improve accuracy but effects are small.
We observed that end-to-end training alone is insufficient to recover from the heuristic initialization.

\noindent \textbf{Stability of End-to-End Training.}
For three out of six Cambridge scenes, DSAC's end-to-end training did not converge despite manually tuning training parameters.
In contrast, end-to-end training of our method converges for all scenes due to learning with broad hypothesis distributions, and our improved approximation of refinement gradients.

\noindent \textbf{Varying Output Resolution.} 
We analyze the impact of the FCN output resolution, \ie the number of scene coordinates predicted, on pose estimation accuracy.
The default output resolution of our FCN architecture is $80 \times 60$.
We simulate smaller output resolutions by sub-sampling correspondences during test time.
We simulate higher output resolutions by executing the FCN multiple times with shifted inputs.
See results in Table \ref{table:7scenesdetails} (right) for the 7Scenes dataset. 
We observe a graceful decrease in accuracy with smaller output resolutions.
Therefore, the runtime of the pipeline could potentially be optimized by predicting less scene coordinates while maintaining high accuracy.
On the other hand, we observe only a small increase in accuracy with higher output resolutions, \ie when the FCN predicts more scene coordinates.
Therefore, we do not expect an advantage in using up-sampling layers to produce full resolution outputs.

\noindent \textbf{Scene Coordinate Initialization.}
When training our pipeline without a 3D scene model, we initialize scene coordinates to have a constant distance $d$ from the camera plane, see Sec.~\ref{sec:meth:training}.
We set this value to $d=3$m for indoor scenes and $d=10$m for outdoor scenes, according to the coarse range of depth values we expect for these settings.
Without this initialization, scene coordinate predictions might lie behind the camera or near the projection center in the beginning of training, resulting in unstable gradients and very low test accuracy.
We found the aforementioned values for $d$ to generalize well on the diverse set of scenes we experimented on.
However, setting $d$ to a value that is far off the actual range of distances can harm accuracy.
For example, when setting $d=10$m for the 7Scenes dataset, test accuracy decreases to 49.3\%.

\begin{figure*}[th!]
\begin{center}
   \includegraphics[width=0.9\linewidth]{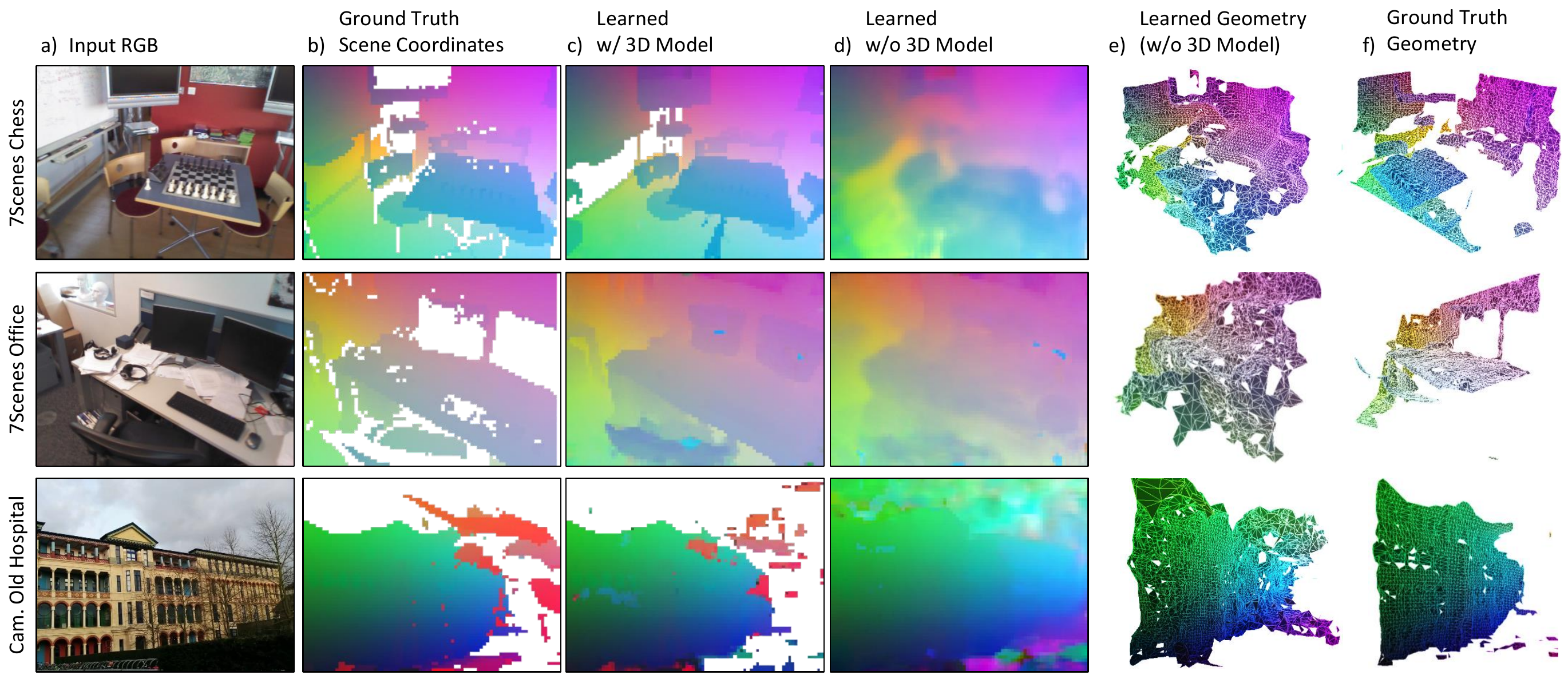}
\end{center}

   \caption{\textbf{Learning Scene Geometry.} We show scene coordinate predictions for test images after end-to-end optimization. For comparison, we calculate ground truth scene coordinates (\textbf{b}) using measured depth (7Scenes) or the 3D scene model (Cambridge Hospital). When trained using a 3D model (\textbf{c}), our method learns the scene geometry accurately. When trained without a 3D model (\textbf{d}), our method discovers an approximate geometry, automatically. The last two columns show a 3D mesh representation of column (d) and (b).}
\label{fig:geometry}
\vspace{-0.3cm}
\end{figure*}

\noindent \textbf{Learning Scene Geometry.}
We visualize the approximate scene geometry discovered by our system when trained without a 3D model in Fig.~\ref{fig:geometry}.
Although our heuristic initialization ignores scene geometry entirely, the system recovers depth information through optimization of reprojection errors.

\noindent \textbf{Run Time.}
The total processing time of our implementation is $\approx$200ms per image on a Tesla K80 GPU and an Intel Xeon E5-E5-2680 v3 CPU (6 cores).
Time is spent mainly for scene coordinate regression, including the overhead of transferring data between the pose estimation front-end (C++) and the deep learning back-end (LUA/Torch), as well as transferring data between main memory and GPU memory.
Pose optimization takes $\approx$10ms on the aforementioned CPU.
Training the pipeline takes 1-2 days per training stage and scene on a single Tesla K80 GPU.

\section{Conclusion}
We have presented a method for 6D camera pose estimation from \rgb images which surpasses accuracy of sparse feature-based and learning-based competitors. 
The system can learn from small training sets, and still generalize well to unseen views.
Training can utilize a 3D scene model, or discover scene geometry automatically.
We make the source code and trained models publicly available\footnote{\url{https://hci.iwr.uni-heidelberg.de/vislearn/research/scene-understanding/pose-estimation/\#CVPR18}}.

Our method scales to large outdoor scenes but fails on city-scale scenes like the challenging Cambridge Street scene \cite{kendall2015convolutional}. 
In the future, we could pair our system with an image retrieval approach as proposed in \cite{sattler2017largescale}.

\paragraph*{Acknowledgements:}
This project has received funding from the European Research Council (ERC) under the European Union’s Horizon 2020 research and innovation programme (grant agreement No 647769). The computations were performed on an HPC Cluster at the Center for Information Services and High Performance Computing (ZIH) at TU Dresden. We thank Tobias Witt for his help visualizing our results, and Lynton Ardizzone for inspiring discussions on the differentiability of iterative optimization.

\appendix

\section{Parameter Listing}
\label{sec:parameters}

We list the main parameter settings for executing (Sec.~\ref{sec:pipeline}) and learning (Sec.~\ref{sec:learning}) our pipeline. 
We use the same parameter settings for all scenes, indoor and outdoor, expect for the scene coordinate initialization parameter $d$.
See details below.

\subsection{Pipeline Parameters}
\label{sec:pipeline}

Our FCN \cite{fcn2015} network architecture takes an image of $640 \times 480$ px as input. 
We re-scale larger images to $480$ px height.
Should an image be wider than $640$ px after re-scaling, we crop it centrally to $640$ px width.
After scene coordinate prediction, we sample $n=256$ hypotheses using random 4-tuples of points and the algorithm of \cite{gao2003complete}.
We reject hypotheses where the reprojection error of the corresponding 4-tuple is larger than the inlier threshold $\tau$, and sample again.
We set the inlier threshold $\tau=10$px for all scenes.
For the soft inlier count (Eq.~\ref{eq:score}) we use a softness factor $\beta=0.5$.
We refine the selected hypothesis until convergence or for a maximum of 100 iterations.

\subsection{Learning Parameters} 
\label{sec:learning}

Our FCN \cite{fcn2015} network architecture predicts one scene coordinate for each $8 \times 8$ px image block.
To make full use of the training data, we randomly shift training images by a maximum of $8$ px, horizontally and vertically. 
We re-scale training images to $480$ px height.
Should an image be wider than $640$ px after re-scaling, we crop it to $640$ px width using random horizontal offsets.

We optimize using ADAM \cite{adam2014} and a batch size of 1 image.
In the following, we state learning hyper-parameters for the three different training steps.

\noindent \textbf{Scene Coordinate Initialization.}
We use an initial learning rate of $10^{-4}$, and train for 300k iterations.
After 100k iterations we halve the learning rate every 50k iterations.

When initializing the pipeline using our scene coordinate heuristic instead of rendered scene coordinates, we reduce training to 100k iterations and utilize only $5\%$ of the training data. 
This is to avoid overfitting to the heuristic.
For the heuristic, we use a constant depth prior of $d=3$m for indoor scenes, and $d=10$m for outdoor scenes.

\noindent \textbf{Optimizing Reprojection Error.}
We train for 300k iterations with an initial learning rate of $10^{-4}$.
After 100k iterations we halve the learning rate every 50k iterations.~We clamp gradients to $\pm 0.5$ before passing them to the FCN.

\noindent \textbf{End-to-End Optimization.}
We use an initial learning rate of $10^{-6}$, and train for 50k iterations.
We halve the learning rate after 25k iterations.
We clamp gradients to $\pm 10^{-3}$ before passing them to the FCN.

For our entropy control schema, we set scale parameter $\alpha=0.1$, initially. 
We optimize using ADAM \cite{adam2014} and a learning rate of $10^{-3}$ for a target entropy of $S^*=6$ bit.

{\small
\bibliographystyle{ieee}
\bibliography{dsac++}

\begin{thebibliography}{10}\itemsep=-1pt

\bibitem{netvlad2016}
R.~Arandjelovi\'c, P.~Gronat, A.~Torii, T.~Pajdla, and J.~Sivic.
\newblock {NetVLAD}: {CNN} architecture for weakly supervised place
  recognition.
\newblock In {\em CVPR}, 2016.

\bibitem{brachmann2017dsac}
E.~Brachmann, A.~Krull, S.~Nowozin, J.~Shotton, F.~Michel, S.~Gumhold, and
  C.~Rother.
\newblock {DSAC}-{Differentiable RANSAC} for camera localization.
\newblock In {\em CVPR}, 2017.

\bibitem{brachmann2016}
E.~Brachmann, F.~Michel, A.~Krull, M.~Y. Yang, S.~Gumhold, and C.~Rother.
\newblock Uncertainty-driven 6{D} pose estimation of objects and scenes from a
  single {RGB} image.
\newblock In {\em CVPR}, 2016.

\bibitem{cao2013graph}
S.~Cao and N.~Snavely.
\newblock Graph-based discriminative learning for location recognition.
\newblock In {\em CVPR}, 2013.

\bibitem{cavallari2017fly}
T.~Cavallari, S.~Golodetz, N.~A. Lord, J.~Valentin, L.~Di~Stefano, and P.~H.
  Torr.
\newblock On-the-fly adaptation of regression forests for online camera
  relocalisation.
\newblock In {\em CVPR}, 2017.

\bibitem{ransac1981}
M.~A. Fischler and R.~C. Bolles.
\newblock {Random Sample Consensus}: {A} paradigm for model fitting with
  applications to image analysis and automated cartography.
\newblock {\em Commun. ACM}, 1981.

\bibitem{Foerstner2016Photogrammetric}
W.~F{\"o}rstner and B.~P. Wrobel.
\newblock {\em {Photogrammetric Computer Vision -- Statistics, Geometry,
  Orientation and Reconstruction}}.
\newblock 2016.

\bibitem{gao2003complete}
X.-S. Gao, X.-R. Hou, J.~Tang, and H.-F. Cheng.
\newblock Complete solution classification for the perspective-three-point
  problem.
\newblock {\em TPAMI}, 2003.

\bibitem{guzman2014multi}
A.~Guzman-Rivera, P.~Kohli, B.~Glocker, J.~Shotton, T.~Sharp, A.~Fitzgibbon,
  and S.~Izadi.
\newblock Multi-output learning for camera relocalization.
\newblock In {\em CVPR}, 2014.

\bibitem{geometricloss}
A.~Kendall and R.~Cipolla.
\newblock Geometric loss functions for camera pose regression with deep
  learning.
\newblock In {\em CVPR}, 2017.

\bibitem{kendall2015convolutional}
A.~Kendall, M.~Grimes, and R.~Cipolla.
\newblock {PoseNet}: {A} convolutional network for real-time 6-{DoF} camera
  relocalization.
\newblock In {\em ICCV}, 2015.

\bibitem{adam2014}
D.~P. Kingma and J.~Ba.
\newblock Adam: {A} method for stochastic optimization.
\newblock {\em CoRR}, 2014.

\bibitem{li2010location}
Y.~Li, N.~Snavely, and D.~P. Huttenlocher.
\newblock Location recognition using prioritized feature matching.
\newblock In {\em ECCV}, 2010.

\bibitem{li2016worldwide}
Y.~Li, N.~Snavely, D.~P. Huttenlocher, and P.~Fua.
\newblock Worldwide pose estimation using 3{D} point clouds.
\newblock In {\em ECCV}. 2012.

\bibitem{lim2012real}
H.~Lim, S.~N. Sinha, M.~F. Cohen, and M.~Uyttendaele.
\newblock Real-time image-based 6-dof localization in large-scale environments.
\newblock In {\em CVPR}, 2012.

\bibitem{fcn2015}
J.~Long, E.~Shelhamer, and T.~Darrell.
\newblock Fully convolutional networks for semantic segmentation.
\newblock In {\em CVPR}, 2015.

\bibitem{rfvscnn2016}
D.~Massiceti, A.~Krull, E.~Brachmann, C.~Rother, and P.~H.~S. Torr.
\newblock Random forests versus neural networks - what's best for camera
  localization?
\newblock In {\em ICRA}, 2017.

\bibitem{sattler2015hyperpoints}
T.~Sattler, M.~Havlena, F.~Radenovic, K.~Schindler, and M.~Pollefeys.
\newblock Hyperpoints and fine vocabularies for large-scale location
  recognition.
\newblock In {\em ICCV}, 2015.

\bibitem{sattler2016large}
T.~Sattler, M.~Havlena, K.~Schindler, and M.~Pollefeys.
\newblock Large-scale location recognition and the geometric burstiness
  problem.
\newblock In {\em CVPR}, 2016.

\bibitem{sattler2016efficient}
T.~Sattler, B.~Leibe, and L.~Kobbelt.
\newblock Efficient \& effective prioritized matching for large-scale
  image-based localization.
\newblock {\em TPAMI}, 2016.

\bibitem{sattler2017largescale}
T.~Sattler, A.~Torii, J.~Sivic, M.~Pollefeys, H.~Taira, M.~Okutomi, and
  T.~Pajdla.
\newblock {Are Large-Scale 3D Models Really Necessary for Accurate Visual
  Localization?}
\newblock In {\em {CVPR}}, 2017.

\bibitem{schindler2007city}
G.~Schindler, M.~Brown, and R.~Szeliski.
\newblock City-scale location recognition.
\newblock In {\em CVPR}, 2007.

\bibitem{shotton13scorf}
J.~Shotton, B.~Glocker, C.~Zach, S.~Izadi, A.~Criminisi, and A.~Fitzgibbon.
\newblock Scene coordinate regression forests for camera relocalization in
  {RGB-D} images.
\newblock In {\em CVPR}, 2013.

\bibitem{Simonyan2014}
K.~Simonyan and A.~Zisserman.
\newblock Very deep convolutional networks for large-scale image recognition.
\newblock {\em CoRR}, 2014.

\bibitem{svarm2014accurate}
L.~Svarm, O.~Enqvist, M.~Oskarsson, and F.~Kahl.
\newblock Accurate localization and pose estimation for large 3{D} models.
\newblock In {\em CVPR}, 2014.

\bibitem{mlesac2000}
P.~H.~S. Torr and A.~Zisserman.
\newblock {MLESAC}: {A} new robust estimator with application to estimating
  image geometry.
\newblock {\em CVIU}, 2000.

\bibitem{valentin2016learning}
J.~Valentin, A.~Dai, M.~Nie{\ss}ner, P.~Kohli, P.~Torr, S.~Izadi, and
  C.~Keskin.
\newblock Learning to navigate the energy landscape.
\newblock {\em CoRR}, 2016.

\bibitem{valentin2015cvpr}
J.~Valentin, M.~Nie{\ss}ner, J.~Shotton, A.~Fitzgibbon, S.~Izadi, and P.~H.~S.
  Torr.
\newblock Exploiting uncertainty in regression forests for accurate camera
  relocalization.
\newblock In {\em CVPR}, 2015.

\bibitem{LSTMPoseNet}
F.~Walch, C.~Hazirbas, L.~Leal{-}Taix{\'{e}}, T.~Sattler, S.~Hilsenbeck, and
  D.~Cremers.
\newblock Image-based localization with spatial {LSTMs}.
\newblock In {\em ICCV}, 2017.

\end{thebibliography}
}

\end{document}